\documentclass[runningheads]{llncs}
\usepackage{graphicx}
\usepackage{amsmath}
\usepackage{cite}
\usepackage{subfigure}
\usepackage{multirow}
\usepackage{array}
\usepackage{amssymb}
\usepackage{pifont}
\usepackage{textcomp}
\usepackage{CJKutf8}

\newcommand{\xmark}{\ding{55}}
\newcommand*\samethanks[1][\value{footnote}]{\footnotemark[#1]}
\newcommand{\fig}[1]{Fig. \ref{#1}}
\newcommand{\eq}[1]{Eq.\ref{#1}}
\newcommand{\tab}[1]{Table \ref{#1}}

\begin{document}
\begin{CJK}{UTF8}{gbsn}
\title{Maximum Entropy Regularization and\\Chinese Text Recognition}
\titlerunning{MER and Chinese Text Recognition}
%
\author{Changxu Cheng \thanks{Equal contribution.} \and
Wuheng Xu \samethanks[1] \and
Xiang Bai \and 
Bin Feng  \and
Wenyu Liu}
\authorrunning{Cheng, C., Xu, W. et al.}
%
\institute{Huazhong University of Science and Technology\\
\email{\{cxcheng, xwheng, xbai, fengbin, liuwy\}@hust.edu.cn}}
\maketitle              
\begin{abstract}
Chinese text recognition is more challenging than Latin text due to the large amount of fine-grained Chinese characters and the great imbalance over classes, which causes a serious overfitting problem. We propose to apply Maximum Entropy Regularization to regularize the training process, which is to simply add a negative entropy term to the canonical cross-entropy loss without any additional parameters and modification of a model. We theoretically give the convergence probability distribution and analyze how the regularization influence the learning process. Experiments on Chinese character recognition, Chinese text line recognition and fine-grained image classification achieve consistent improvement, proving that the regularization is beneficial to generalization and robustness of a recognition model.

\end{abstract}
\section{Introduction}
Text recognition is a popular topic in the deep learning community. Most of the existing deep learning-based works~\cite{shi2016end,shi2018aster,lyu20192d,liao2019scene,puigcerver2017multidimensional,reeve2019scalable} pay attention to Latin script and achieve good performance.

\begin{figure}[htbp]
    \centering
    \subfigure[]{
        \centering
        \includegraphics[width=0.25\textwidth]{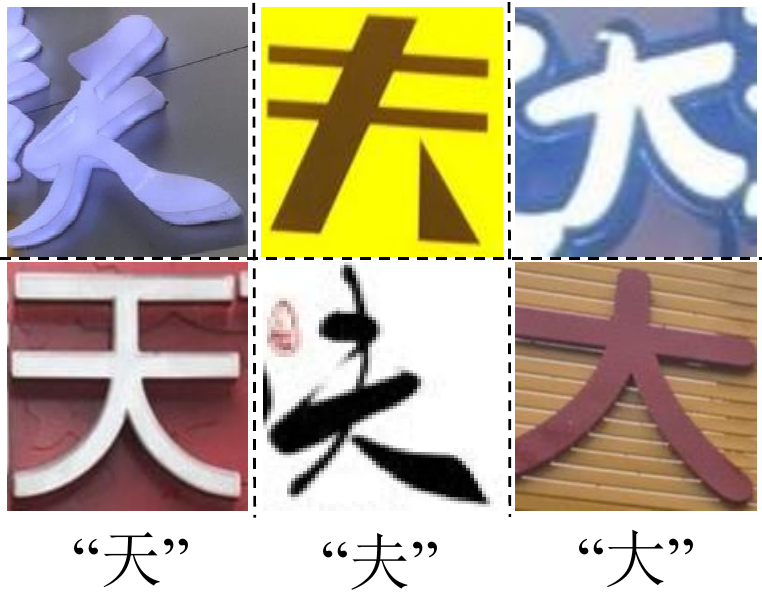}
        \label{fg}
    }
    \quad
    \subfigure[]{
        \centering
        \includegraphics[width=0.65\textwidth]{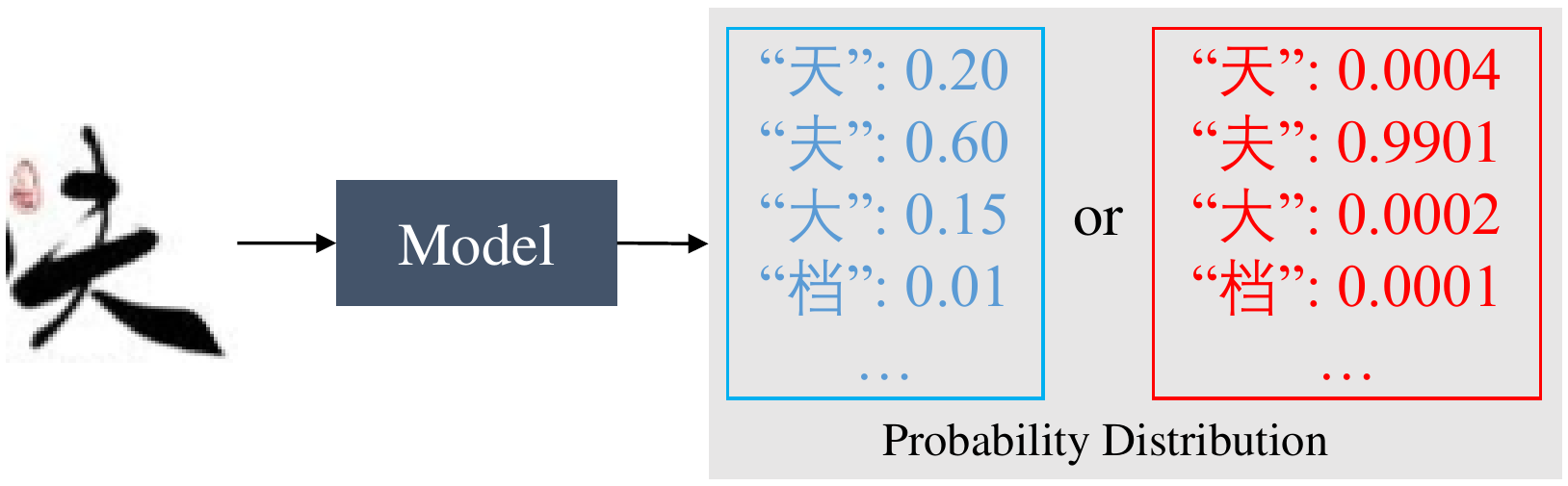}
        \label{pd}
    }
    \caption{Example of Chinese characters and 2 kinds of model prediction. (a) The 2 images in each column are a same character class and the 3 in each row are different. It shows the high inter-class similarity and the large intra-class variance which is a fine-grained attribute. (b) Although the 2 different probability distributions have a same prediction in training set, the left one holds a higher entropy that can describe the learned feature better. Obviously, ``天" is far more similar to ``夫" than ``档", so the confidences on them is supposed to have a large distinction.}\label{intro}
\end{figure}

However, Chinese text differs much from Latin text. There are thousands of common Chinese characters appearing as various styles. Chinese text recognition can be regarded as a kind of fine-grained image classification due to the high inter-class similarity and the large intra-class variance, as shown in \fig{fg}. Besides, there is usually a great data imbalance over character classes\cite{yuan2018chinese,yuan2019large}. These features cause a large demand for training data. Thus character-based recognition model is prone to overfit. Radical-based methods\cite{zhang2018radical,ma2008new,wang2017radical,wu2019joint} convert Chinese characters to radicals as the basic class to simplify the character structure and decrease the number of classes, thus reducing the demand for training data. They are mentioned here to prove the large demand for data if we use character-based recognition. Nevertheless, they are not flexible enough for various circumstances, e.g., handwritten text\cite{zhang2019deep}. And current methods in that way cost much time in practice due to the use of RNN-based decoder.

A common practice to deal with the overfitting problem is to regularize the model during training. There are several techniques aiming at this, including dropout\cite{srivastava2014dropout}, L2 regularization, batch normalization~\cite{ioffe2015batch}. They act on model parameters or hidden units like a blackbox. 
We consider regularization from the perspective of entropy. 
The predicted probability distribution are an indication of how the network generalizes\cite{pereyra2017regularizing}. In Chinese character recognition, we hope that a similar negative class is assigned a larger probability than the dissimilar one given an input image. It requires the probability of positive class to be not that large to leave probability space for other negative classes, which causes a big entropy. This is okay bacause we can recognize correctly as long as the probability of the positive class is the largest. The maximum entropy principle~\cite{jaynes1957information} also points out that the model with the largest entropy of output distribution can represent features best. \fig{pd} illustrates the idea by comparing the 2 probability distributions with high and low entropy respectively. Hence we adopt Maximum Entropy Regularization\cite{pereyra2017regularizing} to regularize the training process.

In this paper, we perform an in-depth analysis of Maximum Entropy Regularization theoretically. The cross-entropy loss and the negative entropy term behave like two 1-dimensional forces which function on the output probability distribution. The elegant gradient function illustrates the regularized behaviour from the perspective of backward propagation. Under an assumption similar to label smoothing, we formulate the convergence probability distribution in training set, which is exactly a relationship between the convergence probability of positive class and the coefficience. 

We conduct experiments on Chinese character recognition, Chinese text line recognition and fine-grained image classification and gain consistent improvement. In addition, we find that model trained with MER can attend on more compact and discriminative regions and filter much noisy area. MER also makes model more robust when label corruption is exerted to our training data.

\section{Related Works}
Our work focuses on model regularization and its application to Chinese text recognition. Here we briefly review some recent works about these two aspects.

\subsection{Model Regularization}
Large deep neural network is prone to overfit in many cases. To relieve the problem, model regularization is commonly used during training.
Dropout~\cite{srivastava2014dropout} is to randomly drop some neurons or connections with a certain probability in layers.
L2 regularization is also called weight decay, which restricts the magnitude of model weights.
Batch normalization~\cite{ioffe2015batch} normalizes hidden units in a training batch to reduce internal covariate shift.
These methods act on model parameters or hidden layers, which is hard to control and not intuitive.
Mixup~\cite{zhang2017mixup} simply uses linear operation on both input images and their labels with an assumption that linear interpolations of features should lead to linear interpolations of the associated targets.

Recently, output distribution of neural network has earned much attention. 
Knowledge distillation~\cite{hinton2015distilling} is proposed to train a small-size model to have a similar output distribution to a large model since the ``soft targets" can transfer the generalization ability, which indicates the effect of output distribution.
Label smoothing~\cite{szegedy2016rethinking} is proposed to encourage the prediction to be less confident by disturbing the one-hot ground truth label with a uniform distribution, which actually adds a KL-divergence (between the uniform and the output distribution) term to the cross-entropy loss in the view of loss function.
Label smoothing is able to improve generalization and model calibration, and benefit distillation~\cite{muller2019does}.
Softermax loss~\cite{pa2019} prompts the summation of top-$k$ output probabilities to be as great as possible to alleviate the extreme confidence caused by cross-entropy loss. But the hyperparameter $k$ is hard to choose.
Bootstrapping~\cite{reed2014training} aims at leveraging ground truth distribution and output distribution as expected distribution, thus the model can train well on dataset with noisy label~\cite{frenay2013classification}. But the output entropy is still as low as that with cross-entropy loss.
According to maximum entropy principle~\cite{jaynes1957information}, the model whose probability distribution represents the current state of knowledge best is the one with the largest entropy. 
Correspondingly, a maximum-entropy based method, called confidence penalty~\cite{pereyra2017regularizing}, includes a negative entropy term in loss function, which acts similarly to label smoothing but performs better. 
As a result, it has attracted researcher's interest in applying it to multiple tasks, e.g., sequence modeling~\cite{liu2018connectionist}, named entity recognition~\cite{yepes2018confidence} and fine-grained image classification\cite{mefg}.
CTC~\cite{graves2006connectionist}-based sequence model trys to penalizes peaky distributions by using maximum-entropy based regularization~\cite{liu2018connectionist}.
However, none of these works make a deep analysis on the regularization term or the whole loss function theoretically or experimentally.

\subsection{Chinese Text Recognition}
Chinese text recognition is more challenging than Latin script due to the more character categories and the more complicate layout. There are usually two tasks derived from Chinese text recognition: Chinese character recognition (CCR) and Chinese text line recognition (CLR).

As for Chinese character, recent methods can be devided into two streams: character-based CCR (CCCR)\cite{zhong2015multi} and radical-based CCR (RCCR)\cite{zhang2018radical}. 
Taken as a single class, every Chinese character is well classified with deep learning\cite{zhang2017online}. However, CCCR has no capability to handle unseen characters. 
By considering the structure of Chinese character, RCCR methods exploit radicals to represent a character\cite{ma2008new, wang2017radical, zhang2018radical, wu2019joint}. 
Multi-label learning has been used to detect radicals\cite{wang2017radical}. 
Radical analysis network (RAN)\cite{zhang2018radical} takes the spatial structure of a single Chinese character as a radical sequence and decodes with an attention-based RNN. 
JSRAN\cite{wu2019joint} improves RAN by jointly using STN\cite{jaderberg2015spatial} and RAN. 
However, RAN is very time-consuming during inference, and RCCR is hard to tackle some nonstandard handwritten text. In this paper, we choose CCCR and use SE-ResNet-50\cite{hu2018squeeze} as backbone for CCR.

As for Chinese text line, currently there are also two streams: one is Convolutional Recurrent Neural Network (CRNN)\cite{shi2016end} with CTC\cite{graves2006connectionist}, the other is attention-based Encoder-Decoder\cite{shi2018aster}. The vanilla version of them can only process image one-dimensionally.
2D-attention\cite{lyu20192d,liao2019scene} decodes an encoded text image from two-dimensional perspective, which is an extension of the latter. 
In our experiments, we simply use 1-d attention-based Encoder-Decoder for CLR.

\section{Analysis on Maximum Entropy Regularization}
Unlike the previous works which only analyze the entropy term, we study both the term and the complete loss function to discover the joint effect.

\subsection{Review of Cross-Entropy Loss}
We first review the common operation in a classification problem.
Given an input sample $x$ with label $y$, a classification model produces $\mathrm{C}$ scores $\left\{z_i \right\}_{i=1}^\mathrm{C}$. Then we canonically get the output probability distribution \textbf{p} by softmax function:
\begin{equation}
    p_i=\frac{e^{z_i}}{\sum \limits_j e^{z_j}}
\end{equation}
The derivative of softmax is:
\begin{equation}
    \label{pz}
    \frac{\partial p_i}{\partial z_j} = 
    \begin{cases}
        p_j \left(1-p_j \right), & \mbox{$i=j$}\\
        -p_i p_j, & \mbox{$i\neq j$}
    \end{cases}
\end{equation}
The cross-entropy (CE) loss and its derivative are:
\begin{equation}
    L_{\mathrm{CE}}=-\log p_y
\end{equation}
\begin{equation}
    \label{grad_ce}
    \frac{\partial L_{\mathrm{CE}}}{\partial z_i} = 
    \begin{cases}
        p_i-1<0, & i=y\\
        p_i>0, & i\neq y
    \end{cases}
\end{equation}

Optimized by using gradient descent, the model prompts the probability of the $y$-th class to be higher and higher and that of other classes to be lower and lower, which leads to a approximation of one-hot vector in training dataset. Consequently we get confident output with low entropy that is often a sympton of overfitting~\cite{szegedy2016rethinking}. In image classification, a confident model tends to focus on many regions, even including noisy background, to have sufficient clues for assertive prediction.

\subsection{Maximum Entropy Regularization}
What really matters is the discriminative region instead of noisy background in training set, which brings more uncertainty of depicting an image of a class. In other words, the prediction has a large entropy.

We would like to regularize the entropy of the output probability distribution $\left\{p_i \right\}_{i=1}^\mathrm{C}$ to make model more general and alleviate the overfitting pain. The entropy is formulated as:
\begin{equation}
    H\left(\textbf{p}\right)=-\sum \limits_{i=1}^{\mathrm{C}}p_i \log p_i
\end{equation}

\begin{figure}[htb]
    \begin{center}
        \includegraphics[scale=0.8]{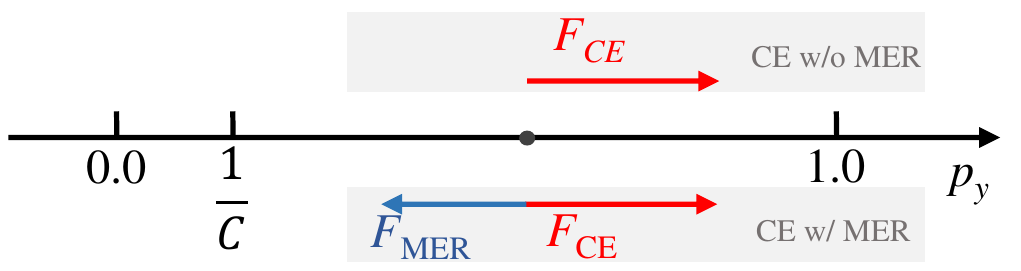}
    \end{center}
    \caption{Illustration of how MER term influences the convergence probability of positive class. Cross-entropy loss without MER always pushes the convergence probability to 1.0, whereas MER term pushes the probabilities to uniform distribution. They behave like two kinds of force, and the probability will finally reach to a point where the 2 forces get balanced.}\label{force}
\end{figure}

Mathematically, the entropy $H\left(\textbf{p}\right)$ reaches the minimum when \textbf{p} is a one-hot vector, and maximum when \textbf{p} is the uniform distribution. The former is realized automatically by the vanilla cross-entropy loss, while the latter is promising to contribute to regularizing. Hence we take the negative entropy as Maximum Entropy Regularization (MER) term which is directly imposed on the common cross-entropy loss function:
\begin{equation}
    L_{\mathrm{MER}}=-H\left(\textbf{p}\right)
\end{equation}
\begin{equation}
    \label{reg0}
    L_{\mathrm{REG}}=L_{\mathrm{CE}}+\lambda L_{\mathrm{MER}}
\end{equation}
where $\lambda$ is the hyperparameter deciding the influence of MER. Intuitively, MER reduces the extreme confidence caused by cross-entropy loss. $L_{\mathrm{CE}}$ and $L_{\mathrm{MER}}$ perform like two kinds of forces that push the probability of positive class to opposite direction, as shown in \fig{force}. The subsequent part illustrates it from the perspective of gradient.

\subsection{Derivative of Regularized Loss}
We now consider the derivative of the regularized loss with respect to output scores $\left\{z_i \right\}_{i=1}^\mathrm{C}$ which is directly related to the model, just like \eq{grad_ce}.

The derivative with respect to the probability distribution is:
\begin{equation}
    \label{grad_0}
    \frac{\partial L_{\mathrm{REG}}}{\partial p_i} = 
    \begin{cases}
        -\frac{1}{p_i}+\lambda \left(\log p_i +1\right), & i=y\\
        \lambda \left(\log p_i + 1\right), & i\neq y
    \end{cases}
\end{equation}

According to \eq{grad_0}, \eq{pz} and the chain rule for derivation, the derivative with respect to the score is:
\begin{equation}
    \label{complicate}
    \frac{\partial L_{\mathrm{REG}}}{\partial z_i}=
    \begin{cases}
        p_i \left(1-\frac{1}{p_i}+\lambda \log p_i - \lambda \sum \limits_j p_j \log p_j\right), & i=y\\
        p_i \left(1 + \lambda \log p_i - \lambda \sum \limits_j p_j \log p_j\right), & i\neq y
    \end{cases}
\end{equation}

By defining a cell function $f$:
\begin{equation}
    f(q)=q\left(1 + \lambda \log q - \lambda \sum \limits_j p_j \log p_j\right)
\end{equation}
we reformulate \eq{complicate} as:
\begin{equation}
    \label{Lz}
    \frac{\partial L_{\mathrm{REG}}}{\partial z_i}=
    \begin{cases}
        f(p_i)-1, & i=y\\
        f(p_i), & i\neq y
    \end{cases}
\end{equation}

Note that \eq{Lz} has the similar elegant format with \eq{grad_ce}. Differently, with $p_i\in [0, 1]$, the gradient in \eq{Lz} is not always positive or negative, so the probabilities are not decreasing to 0 or increasing to 1 under more distributed scores.

\subsection{Convergence Probability Distribution}
Here we give the theoretical convergence probability distribution with a little strong assumption. To simplify the problem, we now only consider the probabilities instead of scores, which means the softmax operation is ignored:
\begin{equation}
    \label{cons}
    \begin{split}
        \text{min } L_{\mathrm{REG}}&=-\log p_y+\lambda \sum \limits_{i=1}^{\mathrm{C}}p_i \log p_i\\
        \text{s.t. } \sum \limits_{i=1}^{\mathrm{C}}p_i&=1
    \end{split}
\end{equation}

\begin{figure}[htb]
    \begin{center}
        \includegraphics[scale=0.6]{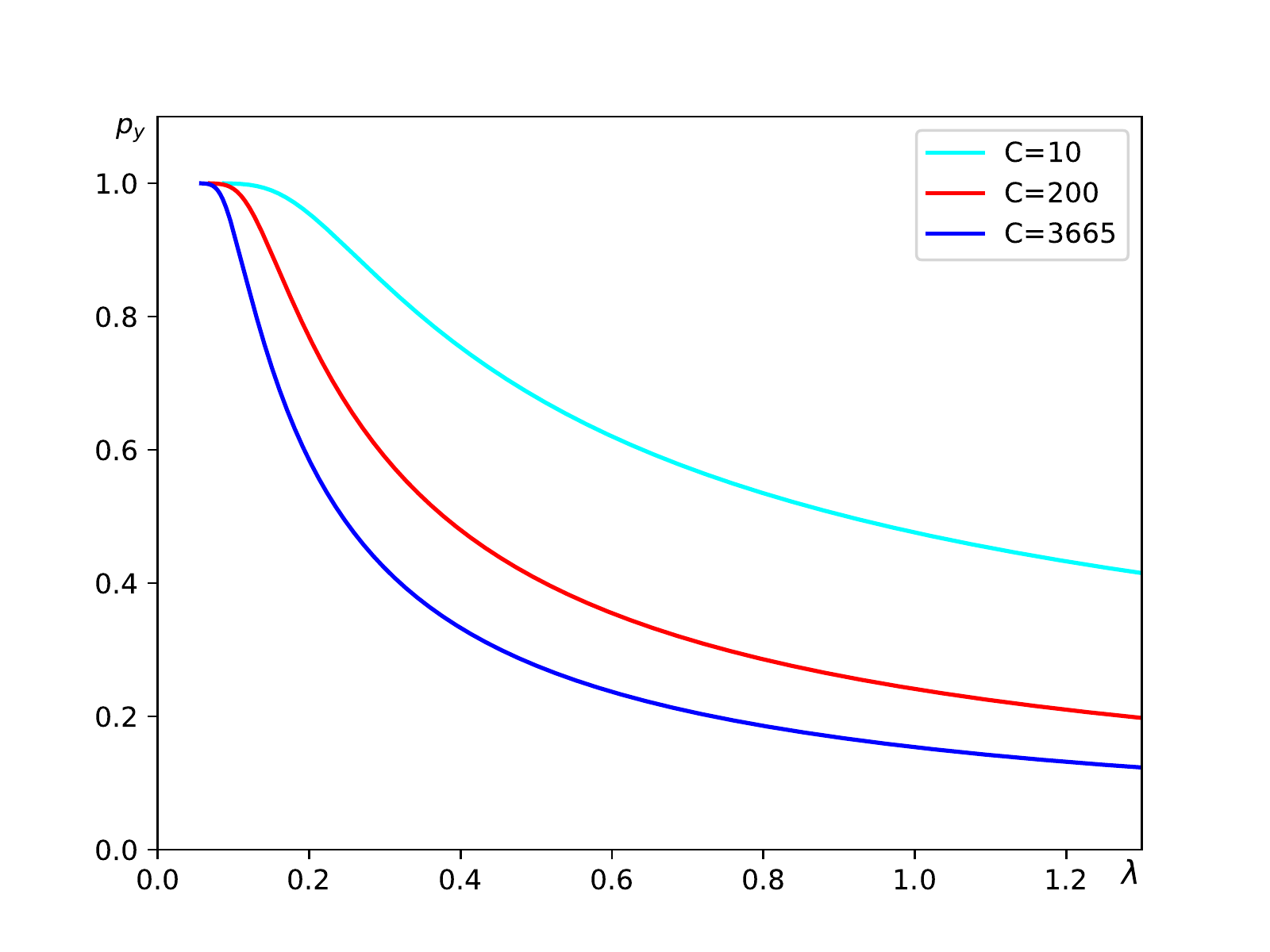}
    \end{center}
    \caption{The $p_y-\lambda$ curve. The convergence probability of positive class decreases as $\lambda$ increases. The more classes we have, the lower probability the model converges to.} \label{curve}
\end{figure}

A natural idea to solve it is to convert it to an unconstrained optimization problem by lagrange multiplier:
\begin{equation}
    \text{min } L_{\mathrm{lag}}=-\log p_y+\lambda \sum \limits_{i=1}^{\mathrm{C}}p_i \log p_i + \alpha \left(\sum \limits_{i=1}^{\mathrm{C}}p_i-1\right)
\end{equation}

We assume that every negative class has the same position. Setting the derivatives with respect to $p_i$ and $\alpha$ to $0$, we get the convergence probability relation between positive class and negative class:
\begin{equation}
    \label{pp}
    p_i=p_y e^{-\frac{1}{\lambda p_y}}, i\neq y
\end{equation}

Futhermore, the relationship between the convergence probability and the coefficience $\lambda$ can be formulated by \eq{pp} and the constrain in \eq{cons} as:
\begin{equation}
    \label{coef}
    \lambda = \frac{m}{\log \frac{\mathrm{C}-1}{m-1}}
\end{equation}
where $m=\frac{1}{p_y}>1$. \eq{pp} and \eq{coef} together indicate the ideal output probability distribution when model converges for a given $\lambda$. To be more intuitive, we plot the $p_y-\lambda$ curve in \fig{curve}. The convergence probability of positive class decreases monotonically with the increase of $\lambda$. When $\lambda \rightarrow+\infty$, $p_y \rightarrow \frac{1}{\mathrm{C}}$, which is exactly a uniform distribution.

Under the assumption, MER has no difference from label smoothing~\cite{szegedy2016rethinking}. They both leverage a uniform distribution to regularize, which is unrealistic for all classes. We can bridge them through the $p_y-\lambda$ curve. Also we get a deeper understanding of $\lambda$ and have some guidance on how to choose a proper $\lambda$.

In fact, MER is stronger than label smoothing since it has more potential beyond the naive assumption. Former~\cite{mefg,pereyra2017regularizing} and present experiments are all illustrating this.

\section{Experiments}
We make several experiments on both Chinese text recognition and fine-grained image classification using PyTorch\cite{paszke2017automatic} to prove the power of MER. Besides, we verify the $p_y-\lambda$ curve, compare MER with label smoothing, and investigate the effectiveness when model is trained with label corruption~\cite{frenay2013classification}.

\subsection{Chinese Text Recognition}
We conduct Chinese character recognition and Chinese text line recognition respectively.

\subsubsection{Datasets}

\paragraph{CTW} is a very large dataset of Chinese text in street view images\cite{yuan2018chinese}. The dataset contains 32,285 images with 1,018,402 Chinese characters from 3850 unique ones. The annotation is only in character level. The texts appear as various styles, including planar text, raised text, text under poor illumination, distant text, partially occluded text, etc.

\paragraph{ReCTS} consists of 25,000 scene text images captured from signboards\cite{ReCTS}. All text lines and characters are labeled. It has 440,027 Chinese characters from 4,435 unique ones and 108,924 Chinese text lines from 4,135 unique characters.

\subsubsection{Implementation Details}
For Chinese character recognition, we take it as a classification problem and use SE-ResNet50\cite{hu2018squeeze} as backbone. Images are resized to the same size $32\times 32$. The training batch size is 128. In CTW, specifically, we use both training and validation set to train with 3665 characters instead of only 1000 common characters~\cite{yuan2019large}.
For Chinese text line recognition, we use attention-based Encoder-Decoder as the same in ASTER\cite{shi2018aster} but without STN framework. Images are resized to $32\times128$. Batch size is 64.

Data augmentation is used, including changing angles in range [-10\textdegree, 10\textdegree], performing perspective transformation and changing the brightness, contrast and saturation randomly. 
We first train a base model from scratch. Then we finetune the pretrained model with/without MER using a same training strategy to keep fair.
Stochastic gradient decent(SGD) with Momentum is used for optimization and the learning rate hits a decay (from 1e-2 to 1e-5, decay rate is 0.1) if the training loss stops falling for a while. We set the weight decay as 1e-4 and momentum as 0.9. 
All the models are trained in one NVIDIA 1080Ti graphics card with 11GB memory. 
It takes less than 12 hours to reach convergence in character recognition, and about 2 days to reach convergence in text line recognition.

\begin{table}[htb]
    \caption{Accuracy of Chinese text recognition}
    \label{tabletext}
    \centering
    \subtable[CTW character recognition]{
        \centering
        \begin{tabular}{|c|c|c|c|c|}
            \hline
            \textbf{Backbone} & ResNet50\cite{yuan2019large} & ResNet152\cite{yuan2019large} & \multicolumn{2}{|c|}{SE-ResNet50}\\
            \hline
            \textbf{MER} & \xmark & \xmark & \xmark & \checkmark\\
            \hline
            \textbf{Acc.} & 78.2 & 79.0 & 78.53 & \textbf{79.24}\\
            \hline
        \end{tabular}\label{ctwres}
    }
    \subtable[Model accuracy with/without MER]{
        \centering
        \begin{tabular}{|c|p{18mm}<{\centering}|c|c|}
            \hline 
            \multirow{2}{*}{\textbf{Method}} 
            & \multicolumn{2}{|c|}{\textbf{Character recognition}} & \textbf{Text line recognition}\\
            \cline{2-4}
                & CTW & ReCTS & ReCTS\\ 
            \hline
            CE w/o MER & 78.53 & 92.44 & 76.89\\ 
            \hline 
            CE w/ MER & \textbf{79.24} & \textbf{92.74} & \textbf{77.50}\\ 
            \hline 
        \end{tabular}\label{text}
    }    
\end{table}

\begin{figure}[htb] 
    \center{\includegraphics[width=4.7in]{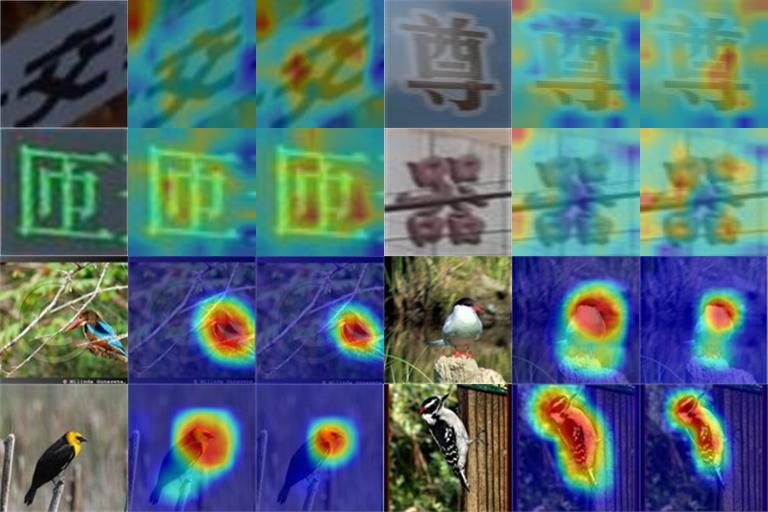}} 
    \begin{flushleft}
    \qquad \textbf{\tiny Input} \quad \qquad \textbf{\tiny CE w/o MER} \quad \textbf{\tiny CE w/ MER} \quad \quad \textbf{\tiny Input} \quad \qquad \textbf{\tiny CE w/o MER} \quad \textbf{\tiny CE w/ MER} 
    \end{flushleft}
    \caption{Visualization of activation maps. Row 1-2 presents Chinese characters, and Row 3-4 presents fine-grained birds. Every triplet contains input image and attention map from model trained with/without MER. Training without MER is prone to make the model focus on broader region, including much background noise. MER regularizes the model to focus on more compact and discriminative region. Note that Row 1-2 is visualized by summing over intermediate feature map channels, whereas Row 3-4 is the class activation map.} \label{visual}
\end{figure}

\subsubsection{Results}
Training using MER improves accuracy without any additional overhead. A model thus can be strengthened easily.
As shown in \tab{tabletext}, model trained with MER outperforms the one without MER on both character recognition and text line recognition. MER can even make a model outperforms another deeper one, like SE-ResNet50 and ResNet152 in \tab{ctwres}.

To be more intuitive, we visualize the region response of character images in CTW by summing over the intermediate feature map channels and exerting min-max normalization. As shown in the top two rows of \fig{visual}, the model trained without MER usually has distributed response and is prone to focus on noisy regions. By using MER, the model concentrates mainly on the text body, thus is more robust to noisy background.

\subsection{Fine-grained Image Classification}
Text recognition can be regarded as a kind of fine-grained image classification since many characters have subtle inter-class but big intra-class difference. To be more general, we also verify the effectiveness of MER on the classical fine-grained image dataset: CUB-200-2011. 

\paragraph{CUB-200-2011} contains 11,788 images of 200 types of birds, 5,994 for training and 5,794 for testing\cite{wah2011caltech}. It is a typical and popular dataset for fine-grained image classification.

\subsubsection{Implementation Details}
When preprocessing the training data, we adopt random crop and random horizontal flip to augment data. Then the images are resized to $448\times 448$. ResNet50\cite{he2016deep} is the backbone network whose parameters are initialized from pretrained model on ImageNet. We train the model for 80 epochs with batch size set as 8 using the Momentum SGD optimizer. Learning rate starts from 1e-3 and decays by $0.3$ when the current epoch is in $[40, 60, 70]$. When we use MER, $\lambda$ is set to $[1.0, 0.5, 0.2, 0.1]$ empirically when the current epoch is $[30, 50, 70]$ respectively.
\subsubsection{Results}
Our simple ResNet50 trained with MER also gains a lot, even outperforms the former complicated models, as shown in \tab{bird}. Hence our model is both accurate and fast.

We visualize the activation map using the last convolutional feature map and the last linear layer weights by CAM\cite{zhou2016learning}. As shown in the bottom two rows of \fig{visual}, MER makes a model focus on more compact and discriminative region, and ignore the noisy background which is harmful to model generalization. In some circumstances, appearance in background or common body region (not discriminative) can make a model more confident on training set, but in this way the really discriminative region are not attended enough.

\begin{table}[htb]
    \centering
    \caption{Accuracy on CUB-200-2011}\label{bird}
    \begin{tabular}{|c|c|c|c|c|c|}
        \hline
        \textbf{Method} & RACNN\cite{fu2017look} & MACNN\cite{zheng2017learning} & MAMC\cite{sun2018multi} & \multicolumn{2}{|c|}{ResNet50}\\
        \hline
        \textbf{MER} & \xmark & \xmark & \xmark & \xmark & \checkmark\\
        \hline
        \textbf{Acc.} & 85.3 & 86.5 & 86.5 & 86.4 & \textbf{87.3}\\
        \hline
    \end{tabular}
\end{table}

\begin{table}[htb]
    \centering
    \caption{Theoretical and experimental convergence probability of positive class (CPP) with different $\lambda$ on CTW character recognition.}
    \label{verify}
    \setlength{\tabcolsep}{8mm}{
    \begin{tabular}{|c|c|c|c|}
    \hline 
    \multirow{2}{*}{\textbf{$\lambda$}} & \multicolumn{2}{|c|}{\textbf{Experimental}} & \textbf{Theoretical}\\
    \cline{2-4}
    & \textbf{CE loss} & \textbf{CPP/\%} & \textbf{CPP/\%}\\
    \hline 
    0.06 & 0.061 & 94.08 & 99.99\\ 
    \hline 
    0.1 & 0.099 & 90.57 & 93.00\\ 
    \hline 
    0.2 & 0.593 & 55.26 & 58.50\\ 
    \hline 
    0.7 & 1.691 & 18.43 & 20.80\\ 
    \hline 
    \end{tabular}}
\end{table}

\subsection{Verification of $p_y-\lambda$ Curve}
We experimentally verify the theoretical convergence probability distribution described as \eq{coef} and \fig{curve}. 

Training with MER on CTW character recognition, we set a fixed $\lambda$ for every experiment. When the model converges, the value of cross-entropy loss in training set is used to calculate the expected experimental convergence probability of positive class (CPP):
\begin{equation}
    p_y=e^{-L_{\mathrm{CE}}}
\end{equation}

The theoretical convergence probability can be got directly from the curve in \fig{curve}.
As shown in \tab{verify}, the experimental value is always slightly lower than the theoretical value. This is normal since there are no perfect models that can fit a complicated distribution completely. What is more, the curve is also only a proximation under an assumption that very negative class has the same position. So they are actually in accordance. As a result, the theoretical curve can be used to estimate the convergence probability distribution in training set roughly, which gives us a practical meaning of $\lambda$ and may guide us to choose a proper $\lambda$.

\subsection{Comparison with Label Smoothing}
Since MER is very similar to label smoothing (LS), we compare their results on CUB-200-2011. LS also has a coefficience $\lambda$ which means the final convergence probability of positive class (CPP) is $\left(1-\lambda+\frac{\lambda}{\mathrm{C}}\right)$. Hence both the two methods have the attribute that CPP decreases as $\lambda$ increases.

Inspired by the theoretical relationship between CPP ($p_y$) and $\lambda$, we choose 4 theoretical CPPs to have 4 pairs of experiments. For each CPP, $\lambda$ of LS is $1-p_y$, and $\lambda$ of MER is chosen from the curve in \fig{curve} as the previous part.

\begin{table}[htb]
    \centering
    \caption{Comparison of MER and label smoothing on CUB-200-2011}\label{lsr}
    \setlength{\tabcolsep}{1.5mm}{
        \begin{tabular}{|c|c|c|c|c|}
        \hline
        \textbf{Theoretical CPP/\%} & \textbf{$\lambda$} & \textbf{Method} & \textbf{Training Entropy} & \textbf{Acc.}\\
        \hline
        \hline
        \multirow{2}{*}{24} & 0.76 & LS & 4.64 & 86.16\\
        \cline{2-5}
        & 1.0 & MER & 4.55 & 87.11\\
        \hline
        \multirow{2}{*}{41} & 0.59 & LS & 3.90 & 86.56\\
        \cline{2-5}
        & 0.5 & MER & 3.71 & \textbf{87.14}\\
        \hline
        \multirow{2}{*}{77} & 0.23 & LS & 1.82 & 87.00\\
        \cline{2-5}
        & 0.2 & MER & 1.35 & 86.73\\
        \hline
        \multirow{2}{*}{90} & 0.10 & LS & 0.92 & 86.61\\
        \cline{2-5}
        & 0.15 & MER & 0.36 & 86.50\\
        \hline
        \hline
        100 & 0.0 & w/o regularization & 0.09 & 86.42\\
        \hline
        \end{tabular}}
\end{table}

\begin{table}[htb]
    \centering
    \caption{Results of model trained with label corruption. Note that $\lambda = 0.0$ means the model is trained without MER.}\label{lc}
    \subtable[CUB-200-2011]{
        \centering
        \begin{tabular}{|c|c|c|c|}
            \hline
            \multirow{2}{*}{\textbf{Corruption Rate}} & \multicolumn{3}{|c|}{\textbf{$\lambda$}}\\
            \cline{2-4}
            & 0.0 & 0.5 & 1.0\\
            \hline
            0.1 & 82.91 & 84.19 & 84.74\\
            \hline
            0.2 & 79.01 & 81.33 & 81.91\\
            \hline
        \end{tabular}
    }
    \subtable[CTW character recognition (Validation Set)]{
        \centering
        \begin{tabular}{|c|c|c|c|}
            \hline
            \multirow{2}{*}{\textbf{Corruption Rate}} & \multicolumn{3}{|c|}{\textbf{$\lambda$}}\\
            \cline{2-4}
            & 0.0 & 0.3 & 0.6\\
            \hline
            0.1 & 84.12 & 84.61 & 84.58\\
            \hline
            0.2 & 82.63 & 83.20 & 83.46\\
            \hline
        \end{tabular}
    }
\end{table}

As shown in \tab{lsr}, MER always gets improvement more or less, while LS is more sensitive to $\lambda$ and can even be harmful (see CPP=24). MER can achieve better accuracy than LS with their own $\lambda$s. Besides, LS only works well when $\lambda$ is small. We argue that the negative influence of the dependence on uniform distribution can be zoomed and nonnegligible with a large $\lambda$ (a small CPP), which limits the potential ability of LS. By contrast, MER is more flexible to regularize the expected probability distribution. With a same theoretical CPP, training entropy of LS is higher than MER, which also reflects the influence of uniform distribution.

\subsection{Train with Label Corruption}
To explore the power of MER on noisy dataset, we randomly corrupt labels of a certain proportion of training images.
Label corruption is usually more harmful than feature corruption\cite{frenay2013classification}. The false labels can mislead the learning process.

We find that the model trained with MER is more robust on both Chinese character recognition and fine-grained classification. In \tab{lc}, corruption rate is the proportion of training images whose labels are randomly corrupted. The model trained without MER ($\lambda =0.0$) suffers a lot because it is always very confident and reach to blind devotion to the given labels, even though some labels are wrong. MER regularizes a model to be less confident of labels in training set, so the learning process is less disturbed.

\section{Conclusion}
In this paper, we make a deep analysis on Maximum Entropy Regularization, including how MER term influence the convergence probability distribution and the learning process. MER improves generalization and robustness of a model without any additional parameters. We employ MER on both Chinese text recognition and common fine-grained classification to alleviate overfitting, and gain consistent improvement. We hope that our theoretical analysis can be useful for the further study on the regularization.

\subsubsection*{Acknowledgments.}
This work was supported by the National Natural
Science Foundation of China (NSFC, grant No. 61733007).

\bibliographystyle{splncs04}
\bibliography{mer}

\end{CJK}
\end{document}